\documentclass{article}

\PassOptionsToPackage{numbers, compress}{natbib}
\usepackage{nips_2017}
\usepackage[utf8]{inputenc} 
\usepackage[T1]{fontenc}    
\usepackage{hyperref}       
\usepackage{url}            
\usepackage{booktabs}       
\usepackage{amsfonts}       
\usepackage{nicefrac}       
\usepackage{microtype}      

\usepackage{amsmath, amssymb}
\usepackage{graphicx}
\usepackage{multirow}

\newtheorem{theorem}{Theorem}
\let\oldtheorem\theorem
\renewcommand{\theorem}{\oldtheorem\normalfont}
\newtheorem{defn}{Definition}
\let\olddefinition\defn
\renewcommand{\defn}{\olddefinition\normalfont}

\title{An approach to reachability analysis for feed-forward ReLU
  neural networks}

\author{Submission number \#1196
}

\begin{document}

\maketitle

\begin{abstract}
  We study the reachability problem for systems implemented as
  feed-forward neural networks whose activation function is implemented
  via ReLU functions. We draw a correspondence between establishing
  whether some arbitrary output can ever be outputed by a neural system
  and linear problems characterising a neural system of interest. We
  present a methodology to solve cases of practical interest by means of
  a state-of-the-art linear programs solver. We evaluate the technique
  presented by discussing the experimental results obtained by analysing
  reachability properties for a number of benchmarks in the literature.
\end{abstract}

\section{Introduction}\label{sec:intro}
Over the past ten years, there has been growing interest in trying to
verify formally the correctness of AI systems. This has been
compounded by recent public calls for the development of
``responsible'' and ``verifiable AI''~\cite{RusselDeweyTegmark15}.
Indeed, since the development of ever more complex and pervasive AI
systems including autonomous vehicles, the need for higher guarantees
of correctness for the systems has intensified. Formal verification is
one of the techniques used in other areas of Computer Science,
including hardware and automatic flight control systems, to debug
systems and certify their correctness. It is therefore expected that
formal methods will contribute to provide guarantees that AI systems
behave as intended.

In the area of multi-agent systems (MAS) there already has been
considerable activity aimed at verifying MAS formally. In one line
efficient model checkers for finite state MAS against expressive
AI-based specifications, such as those based on epistemic logic, have
been
developed~\cite{Kacprzak+07a,LomuscioQuRaimondi15,GammieMeyden04a}.
Abstraction techniques have also been put forward to verify infinite
state MAS~\cite{LomuscioMichaliszyn16a} and approaches for
parameterised verification for MAS and swarms have been
introduced~\cite{KouvarosLomuscio16a,KouvarosLomuscio15b,KouvarosLomuscio16b}.
In a different strand of work, theorem proving approaches have been
tailored to MAS~\cite{alechina+10a,shapiro+05a}, and techniques for
the direct verification of MAS programs have been put
forward~\cite{Bordini+06a}.

While significant results have been achieved in these lines, their
object of study is a system that is given either via a traditional
programming language or a MAS-oriented programming language.
It is however expected that machine learning technology will provide
the backbone for a wide range of AI applications, including robotics,
autonomous systems, and AI decision making systems. With the few
exceptions discussed below, at present there is no methodology for the
verification of systems based on neural networks. This paper aims to
make a contribution on this topic.

To begin this investigation, we consider feed-forward neural-networks
(FFNNs) only, possibly with several layers, where the activation
function is governed by ReLU functions~\cite{haykin+11a,nair+10a}.  We
consider specifications concerning safety only and, in particular, we
study reachability. The method we present enables us to check whether
a particular output, perhaps representing a bug, is ever produced by a
given neural-network. While the target specifications are
comparatively simpler than present research in MAS and reactive
systems, reachability remains of paramount importance in program
analysis as it enables the identification of simple errors.

The rest of the paper is organised as follows. In Section~2, we fix the
notation on ReLU-based FFNNs and mixed integer linear programs. In
Section~3, we present an encoding of neural-networks in terms of linear programs
and formally relate reachability on FFNNs in terms of a
corresponding linear programming problem. In Section~4, we apply the
technique to a pole-balancing problem from the literature and identify
safety features of the corresponding FFNNs. In Section~5, we evaluate
the scalability of the approach by presenting experimental results
obtained on FFNNs of various sizes. We conclude in Section~6 by
discussing further work.

{\bf Related Work.} As stated above, much of the past and present
research on verification of AI systems concerns the analysis of actual
programs or traditional finite-state models representing AI systems
against temporal or AI-based specifications. While the aims are the
same as those in this paper, the object of study is intrinsically
different. In contrast, much of the literature on FFNNs is concerned
with training and performance and does not address the formal
verification question. Currently techniques used for checking the
correctness of networks rely on test datasets which are probablistic
and thus obviously incomplete. The few exceptions for formal
verification of neural networks that we are aware of are the following.

\cite{kurd+2003a} advocates the use of safety specifications to validate
neural networks. The work here presented on reachability partially
falls within the types 3 and 4 of safety which they discuss.  While
the broad direction of the work is in line with what pursued here, no
actual verification method is discussed.  A method for finding
adversarial inputs for ReLU feed forward networks through the use of
linear programming was proposed in~\cite{bastani+16a}.  However, the
LP encoding proposed there is tailored to adversarial inputs and
cannot be applied to reachability. Moreover, finding adversarial
inputs can be thought of as a special case of reachability where the
input set is constrained with respect to a specific input; thus the
formulation here proposed is more general.
Related to this, a method for finding adversarial inputs using a
layer-by-layer approach and employing SMT solvers was recently
proposed~\cite{huang+16a}. This technique supports any activation
function, not just ReLU as we do here. However, because the focus is
on adversarial inputs, as before, the method seems not immediately
generilisable to solving reachability on feed-forward networks.
A methodology for the analysis of ReLU feed-forward networks,
conducted independently from this research, was made available on
ArXiv some time before this submission~\cite{katz+17a}. While the aims
seem largely in line as those presented here, there is no formal
correspondence presented between reachability analysis and linear
programs as we do here. Moreover, the underlying techniques proposed
are different. While their method is based on SMT-solving, we only
use linear programming here. Linear programming is used
in~\cite{katz+17a} as a comparison against SMT, but there is no
mention of any optimisation on the LP engine. In contrast, we here
focus on an efficient LP translation and handle floating point
operations in an optimised manner. Also the scenarios studied are
different from ours and are not released for a comparison. Judging
from the experimental results presented for the LP comparison, the LP
technique used in~\cite{katz+17a} performs significantly less
efficiently than what we develop here. For example, their analysis is
only shown with up to 300 ReLU constraints whereas our technique is
able to handle more than 500 with ease. An in-depth comparison of the
performance of the two different techniques will not be possible
until all data in~\cite{katz+17a} are released.

\section{Preliminaries}\label{sec:background}
 In what follows we fix the notation on the key concepts to be used in
the rest of the paper.

\textbf{Feed-forward neural networks} (FFNN) are the simplest class of neural
networks \cite{haykin+99a}.  Their main distinguishing feature is the
lack of cycles. \begin{defn}[Feed-Forward Neural Network]\label{fnn}
Let $V$ be a set of vertices, $E \subseteq V \times V$ be a set of
edges, $w: E \to \mathbb{R}$ be a
\textit{edge weight function} and $b: V \to \mathbb{R}$ be a \textit{bias
function}.  A weighted, directed, acyclic graph $N = (V, E, w, b)$ is a
\textit{feed-forward neural network} if the following properties hold:
\begin{enumerate}
  \item For $j = 1\dots n$ there exists an ordered collection of
  ordered sets $L^{(j)} \subseteq V$ known as $\textit{layers}$ such
  that $V = \biguplus_{j = 1}^n L^{(j)}$ and for a vertex $v \in
  L^{(j)}\ j < n$, the set of endpoints for edges originating from $v$
  is equal to $L^{(j + 1)}$ and for a vertex $v \in L^{(n)}$, the set
  of endpoints from $v$ is equal to the empty set.
\item Each  $L^{(j)}, j \geq 2$ is associated with  a function
  $\sigma^{(j)}: \mathbb{R} \to \mathbb{R}$ known as
  the \textit{activation function}.
\end{enumerate}
\end{defn}

In this paper we consider only \textit{ReLU} activation
functions. This function has grown in popularity over the past few
years in feed-forward networks, replacing the sigmoid and tanh
functions; this is due to the improvement in convergence to the
function being approximated when training the network. It is defined
as $\text{ReLU}(x) = \max(x, 0)$.

While the network itself is a weighted graph with bias represented as a
function, we can define the traditional vector and matrix representation of bias
and weights respectively.
\begin{defn}[Bias and weights of layer]\label{fnn-bias}
For each layer $L^{(k)},\ k \geq 2$ of a neural network $N$, we define the
\textit{bias vector} to be the vector $b^{(k)} \in R^{m}$ with $m = |L^{(k)}|$
defined elementwise as: \(b^{(k)}_i = b(L^{(k)}_{i})\). We further define the
\textit{weight matrix} to be the matrix $W^{(k)} \in R^{m \times n}$ with $m =
|L^{(k)}|$ and $n = |L^{(k - 1)}|$ defined elementwise as: \(W^{(k)}_{ij} =
w(L^{(k - 1)}_{j}, L^{(k)}_{i})\).
\end{defn}

The main use of neural networks is as universal function approximators. For the
purposes of this paper, we define the function computed by a network.
\begin{defn}[Function computed by network]\label{fnn-layer-fn}
Let $N$ be a neural network and $2 \leq i \leq k$ with $k$ the number of layers in
$N$. Then, layer $L^{(i)}$ defines a function $f^{(i)}: \mathbb{R}^{m} \to
\mathbb{R}^{n}$ known as a \textit{computed function} where $m = |L^{(i - 1)}|$,
$n = |L^{(i)}|$ and defined as $f^{(i)}(x) = \sigma^{(i)}(W^{(i)} x + b^{(i)})$
Further, N defines a function $f: \mathbb{R}^{m} \to \mathbb{R}^{n}$ known as
the \textit{computed function} where $m = |L^{(1)}|$, $n = |L^{(k)}|$ and
defined as: $f(x) = f^{(k)}(f^{(k-1)}(...(f^{(2)}(x))))$
\end{defn}

\textbf{Linear Programming} is an optimisation technique where the objective
function and constraints are linear.  Efficient algorithms exist to
solve linear programming problems efficiently~\cite{winston+87a}.  For
the purposes of this paper consider mixed integer linear programs;
these are linear programs which contain both real and integer
variables.

\begin{defn}[Mixed Integer Linear Programs]\label{milp}
A function $f(x_1, ..., x_n)$ is said to be \textit{linear} if
for some $c \in \mathbb{R}^N$, we have $f(x_1, ..., x_n) = \sum_{i = 1}^N c_i x_i$.

For any linear function $f(x_1, ..., x_n)$, and any $b \in
\mathbb{R}$, $f(x_1, ..., x_n) = b$, $f(x_1, ..., x_n) \leq b$ and
$f(x_1, ..., x_n) \geq b$ are said  to be \textit{linear constraints}.

A \textit{linear program} (LP) is a mathematical optimisation problem where the
the objective function is linear and the constraints on the variables of the
objective function are linear.

A \textit{mixed integer linear program} is a linear program which
allows for constraints which require variables to be integer, i.e.,
constraints of the type $x_i \in \mathbb{Z}$.
\end{defn}

In this paper we use \textit{Gurobi} linear programming solver
\cite{gurobi+16a}. Gurobi has good performance and can be used on a wide range
of problems~\cite{mittelmann+16a}.

\section{Verifying Reachability for FFNN} \label{sec:algorithms}
In this paper we focus on reachability analysis, a particular aspect
of safety analysis.

In general terms reachability analysis consists on finding whether a
certain state (or set of states) of a system can be reached given a
fixed set of initial states of the system. Reachability analysis is
commonly used to identify bugs in software systems, e.g., whether
mutual exclusion is enforced in concurrent applications \cite{magee+06}.

To apply this concept to neural networks, we treat our fixed set of
initial states to be a fixed set of \textit{input vectors} and we
attempt to find out whether any vector in a set of \textit{output
vectors} can be computed by the network from a vector in the input
set.

\begin{defn}[Reachability for FFNN.]\label{fnn-reach}\label{reach}
Suppose $N$ is an FFNN with computing function $f: \mathbb{R}^{m} \to
\mathbb{R}^{n}$ with $m = |L^{(1)}|$, $n = |L^{(k)}|$, where $k$ is the number of
layers in $N$.

Let $I \subseteq \mathbb{R}^{m}$ and $O \subseteq \mathbb{R}^{n}$. We
say that $O$ is \textit{reachable from $I$} using $N$ if $\exists
y \in O, \exists x \in I, f(x) = y$.
\end{defn}

While Definition~\ref{reach} presents the general case, we here focus
on input and output sets that are representable via a finite number of
linear constraints on $\mathbb{R}^{n}$.
Observe this still enables us to capture a large number of systems
since all linear equalities and disequalities are allowed e.g.
verification of ACAS networks performed on in~\cite{katz+17a}
uses linearly definable input and output sets.

\begin{defn}[Linearly Definable Set.]\label{linear-define}
  Let $S \subseteq \mathbb{R}^{n}$. We say that $S$ is linearly
defineable if there exists a finite set of linear constraints $C_S$
such that $S = \{x \in \mathbb{R}^{n} \mid x \text{ satifies every
constraint in } C_S\}$. We define $C_S$ to be the constraint set of S.
\end{defn}

We now show that estabilishing reachability for a neural network with
ReLU activation functions can be rephrased into solving a
corresponding linear program resulting from the linear encoding of the
neural network in question.

While informal linear encodings for individual neurons have been
proposed in the past~\cite{katz+17a}, the one we present here is a
formal one which operates on a layer by layer approach. Moreover, it
only utilises a single binary variable; as we demonstrate later, this
is important for efficiency in practical applications.

\begin{defn}[Linear Encoding for FFNN.]
Let $N$ be an FFNN and $2 \leq i \leq k$ with $k$ the number of
layers in $N$. Suppose further $x^{(i-1)}$ and $x^{(i)}$ are vectors of real
(LP) variables representing the input and output of layer $i$ respectively and
$\delta^{(i)}$ a vector of binary (LP) variables. Then, the set of
\textit{linear constraints encoding layer} $i$, (with a ReLU
activation function) is defined as:
\begin{align*}
  C_i = \{x^{(i)}_j &\geq W^{(i)}_{j} x^{(i-1)} + b^{(i)}_{j},
   x^{(i)}_j \leq W^{(i)}_{j} x^{(i-1)} + b^{(i)}_{j} + M \delta^{(i)}_j, \\
   x^{(i)}_j &\geq 0,
   x^{(i)}_j \leq M (1 - \delta^{(i)}_j) \mid j = 1...|L^{(i)}|\}
\end{align*}
where $M$ a "sufficiently large" constant.

We can define the set of \textit{linear constraints encoding the network} as $C =
\cup_{i=2}^{k} C_i$.
\label{lin-encoding}
\end{defn}

By means of this encoding, we can reduce reachability analysis to
solving a linear program defined on these constraints.

\begin{defn}[LP encoding reachability for FFNN.]\label{LP-enc}
Let $N$ be an FFNN, $C$ its encoding as per
Definition~\ref{lin-encoding}, and $I \subseteq \mathbb{R}^m$
(respectively, $O \subseteq \mathbb{R}^n$) be a set of linearly
definable network inputs (outputs, respectively).

The \textit{linear program encoding the reachability} of $O$ from $I$
through $N$ is given by the objective function $z = 0$ and constraints
$C_{reach} = C_{in} \cup C \cup C_{out}$, with the sign of the
variables unconstrained, where
\begin{itemize}
\item $C_{in}$ is a constraint set for $I$ defined on the same
variables used in the encoding of the input to the second layer of $N$
(i.e.  $x^{(1)}_j$ for $j = 1...m$), and
\item $C_{out}$ is a constraint set for $O$ defined on the same
variables used in the encoding of the output of the last layer of $N$
(i.e.  $x^{(k)}_j$ for $j = 1...n$).
\end{itemize}
\end{defn}

\begin{theorem}[Equivalence between reachability analysis and
corresponding LP problems.]  Suppose $N$ is an FFNN on linearly
definable input $I$ and output $O$.  Let $L$ be the corresponding
linear problem encoding the reachability of $O$ from $I$ through $N$ (Definition~\ref{LP-enc}).

Then, $O$ is reachable from $I$ through $N$ if and only if the linear
program $L$ has a feasible solution $\textbf{x}$.

\textit{Proof sketch.} $\implies$ We have $\exists x \in I, \exists y \in O$
with $f(x) = y$. Then, for the assignment
of LP variables $x^{(1)}_j \rightarrow x_j$ and $x^{(i)}_j \rightarrow
f^{(i)}(...(f^{(2)}(x_j)))$, we have that the LP is feasible.

$\impliedby$ We have $\textbf{x}$ is a feasible solution for $L$. Let $x_j =
\textbf{x}^{(1)}_j$ and $y_j = \textbf{x}^{(k)}_j$. Then, we have $x \in I, y
\in O$ and $y = f(x)$ by definition of the LP.
\end{theorem}

\section{Verifying a Neural Controller for the Inverted Cart Problem}\label{sec:experiments}
We now use the methodology developed in the previous section to verify
several reachability specifications on the well-studied inverted
pendulum controller problem~\cite{barto+83a}.

\textbf{Inverted pendulum on cart problem (IPCP).}
The system is composed by a cart moving along a frictionless track
with bounds at either ends of the track. Attached to the centre of the
cart through the use of a frictionless and unactuated joint is a
pole. The pole acts as an inverted pendulum.

The problem can be expressed in control terms by using two state
variables and their derivatives~\cite{barto+83a}.
\begin{itemize}
  \item Position of the cart on the track, denoted by $x$ and bounded
  by $\pm 2.4$.
  \item Speed of the cart, denoted by  $\dot{x}$ and unbounded.
  \item Angle of the pendulum (counter-clockwise), denoted by $\theta$
  and bounded by $\pm 15^{\circ}$.
  \item Angular velocity of the pendulum (counter-clockwise), denoted by
  $\dot{\theta}$ and unbounded.
\end{itemize}

The output of the controller at every discrete time step is a signal
for the application of a force of $+10N$ or $-10N$ (where the
direction is aligned with $x$). Intuitively, the aim of the controller
is to balance the pendulum on the cart for as long as possible while,
at the same time, remaining both in the bounds of the track and in the
bounds of the angle of the pendulum.

{\bf A neural network controller for the IPCP.}
Due to its inherent non-linearity, the IPCP traditional controller
methods cannot be used to solve the problem. Reinforcement learning
methods can be used to derive a neural network that can be used as a
controller for the system. To create a neural controller we used the
deep learning library {\sc Keras}~\cite{chollet+15a} combined with the
Q-learning library {\sc keras-rl}~\cite{plappert+16a}.

The precise details of the resulting network and the training data are
not relevant for what follows (the supplementary material reports the
details). After training the resulting deep, feed-forward neural
network obtained can be described as follows:
\begin{itemize}
  \item The input layer consists of 4 input nodes, one for each of the
  variables of the system.
  \item The resulting three hidden layers consist of 16
  nodes; each layer uses the ReLU activation function.
  \item The output layer consists of 2 nodes denoting the "q-value"
  of moving left and right respectively. The higher
  the q-value, the higher is the predicted reward for the action. The output
  layer does not use a ReLU function as is standard for most networks.
\end{itemize}

\textbf{Reachability via Linear Programming.}
Having derived a neural-network controller for the IPCP, we now
proceed to analyse it in terms of reachability properties. While the
theoretical encoding of the problem is presented in Section~3, to
solve the resulting problem via an automatic solver, we need to
address the resulting issues in terms of floating point approximations.

\textit{Floating point arithmetic.}
For the encoding to be correct, the constraints present in the
resulting LP must take into account a safe level of floating point
precision and use tolerances when defining the links between the
layers. Not doing so may render the analysis to be unsound. A system
may be assessed to be safe (i.e., unwanted regions of the output may
be shown to be unreachable); but this could be the result of by
approximations (roundings or truncations) due to the underlying
floating point arithmetic.

To address this issue we use we use "epsilon" terms when encoding the
network. These are terms used to link the layers of the network to
allow for small perturbations when invoking the linear program solver.
In combination with this, we change the objective function to minimise
the sum of these epsilon terms, instead of simply using the constant
$0$.

Formally, for each layer, the constraint set changes as follows:
\begin{align*}
  C_i = \{x^{(i)}_j &\geq W^{(i)}_{j} x^{(i-1)} + b^{(i)}_{j} - \epsilon^{(i)}_j,
   x^{(i)}_j \leq W^{(i)}_{j} x^{(i-1)} + b^{(i)}_{j} + M \delta^{(i)}_j + \epsilon^{(i)}_j, \\
   x^{(i)}_j &\geq 0,
   x^{(i)}_j \leq M (1 - \delta^{(i)}_j) \mid j = 1...|L^{(i)}|\}
\end{align*}
where $\epsilon_j$ are non-negative variables. Correspondingly, when
encoding a neural network as linear program, we change the objective
function to be $z
= \sum_{i=2}^{k} \sum_{j=1}^{|L^{(i)}|} \epsilon^{(i)}_j,$ which we
then aim to minimise.

This entails that we aim to find an exact solution if possible but, if
one exists with a small epsilon sum, we can still accept it if it is
within the tolerance of the underlying floating point arithmetic.  We
do this by adding a further constraint of the form
$\sum_{i=2}^{k} \sum_{j=1}^{|L^{(i)}|} \epsilon^{(i)}_j \leq t,$ where
$t$ is the tolerance term.

In practice, for current computers we can take this to be $1e^{-6}$
which is one order of magnitude larger than the machine epsilon for
32-bit floating point numbers. We adopted this value in our
experiments. However, when binary inputs are used, a larger tolerance
value may be required because of the techniques used by solvers. We
adopted a value of $1e^{-4}$ for these problems.

\textbf{Reachability specifications and results.}
In view of the encoding discussed above we can now proceed to verify
the behaviour of the neural-network trained for the IPCP.
We consider the following specifications where in each case $S$ is a tupe of
form $(x, \dot{x}, \theta, \dot{\theta})$.
\begin{enumerate}
  \item Is it ever the case that $Q(S, 10) \ngeq Q(S, -10) + 100$ where
  $S=(0, 0, -5, 0)$? Intuitively, this says that force of $10N$ has a
  Q-reward of at least 100 units greater than $-10N$ for the given
  state $S$. This expresses the fact the controller attempts (in the
  strongest possible sense within the bounds of the problem) to move
  the cart to the right when the pendulum is leaning to the right and
  all other factors are unimportant.

  \item Is it ever the case that $Q(S, 10) \ngeq Q(S, -10) + 100$ where
  $S \in \{(x, \dot{x}, \theta, \dot{\theta}) \mid \lvert
  x \rvert \leq 0.5, \lvert\dot{x}\rvert \leq 0.2, -5 \leq \theta \leq
  -4, \lvert \dot{\theta} \rvert \leq 0.1\}$? This is the same
  specification as above but to be checked on a larger states of
  configurations.

  \item Is it ever the case that $Q(S, 10) \ngeq Q(S, -10) + 10$ where
  $S \in \{(x, \dot{x}, \theta, \dot{\theta}) \mid x \leq
  -2, \rvert \dot{x} \lvert \leq 0.2, -2 \leq \theta \leq
  -1.5, \lvert \dot{\theta} \rvert \leq 0.25\}$? This represents the
  fact that the controller attempts to move the cart to the right when
  the pendulum is not at risk from falling over but the cart is almost
  out of left hand side track bound.

  \item Is it ever the case that $Q(S, 10) \ngeq Q(S, -10) + 10$ where
  $S \in \{(x, \dot{x}, \theta, \dot{\theta}) \mid x \leq
  -2, \rvert\dot{x}\lvert \leq 0.4, -2 \leq \theta \leq 1,
  0 \leq \dot{\theta} \leq 0.1\}$?  This is a relaxation of the above
  specification to analyse a larger set of configurations.
\end{enumerate}

To analyse the specifications above we compiled the network for the
IPCP into an LP as described above. We then used
Gurobi~\cite{gurobi+16a} to solve the corresponding LP problems. A
solution found by Gurobi corresponds to the fact that there is a
configuration can be found solving the problem. In this case, the
specification can be met by the system for the values found. If a
solution cannot be found, since by Theorem~1 the method is complete,
we conclude that no input in the space analysed can produce the output
checked.

Doing as above we promptly obtained the following results (full
benchmarks are reported in the next section).
\begin{enumerate}
  \item No solution could be found satisfying the constraints on the
  input and output of the network. We conclude that our synthesised
  controller does strongly prefer to apply $10N$ to balance the
  pendulum in those circumstances.
  \item As above no solution could be found. Again, we conclude that
  in all the region explored the behaviour of the synthesised
  controller is correct.
  \item Again, no solution could be found satisfying both the inputs
  and the outputs. This indicates that the controller attempts to
  return the cart to the centre of the track when it is one side of
  the track within the range of parameters above.

\item The solver reported the solution $x =
  (-2.0, -0.4, -0.15, 0.1)$ (approximation shown) for the problem
  above. This shows that the controller applies the force in what is,
  intuitively, the incorrect direction when the configuration of the
  system is as above. Note also that in this situation the angle of
  the pendulum would also suggest an application of the force in the
  positive direction.
\end{enumerate}

The analysis conducted above shows that the synthesised controller
does not satisfy our specifications as put forward. The values found
by the solver can be fed to the neural network to confirm the
result. We have effectively found a ``bug'' in the synthesised
controller by using a formal encoding into an LP problem.

We stress the importance of this result which enabled us to find an
error in the resulting network in under 1s. Comparable techniques,
e.g., testing are in incomplete and may take considerably longer to
identify the need for further training.


\section{Experimental Results}\label{sec:benchmarks}
We now report on the experimental results obtained by using the
technique presented in the previous sections on a number of
feed-forward networks.  The experiments were run on an Intel Core
i7-4790 CPU (3.600GHz, 8 cores) running Linux kernel 4.4 upon which we
invoked Gurobi version 7.0. The benchmarks are shown only to evaluate
the scalability of the approach, not as validation for the
corresponding problems. Indeed, for some of the problems below
(Reuters and MNIST), reachability analysis is not
applicable. Moreover, since, as discussed in the Introduction, no
other approach exists for reachability analysis, we are unable to
compare our results to other techniques.

More details on all of these experiments as well as the networks and
the code used to perform verification can be found in the
supplementary material of this paper.

\begin{table}[htb]
\centering
\begin{tabular}{ | c | c | c | c | }
  \hline
  Problem & Layer Sizes & Vars (Continuous, Binary) & Time (s) \\
  \hline
  Inv. Pen. 1 & \multirow{4}{*}{4, 16, 16, 16, 2} & 108, 31 & >0.01 \\
  \cline{1-1}
  \cline{3-3}
  Inv. Pen. 2 & & 140, 39 & 0.02 \\
  \cline{1-1}
  \cline{3-3}
  Inv. Pen. 3 & & 142, 41 & 0.03 \\
  \cline{1-1}
  \cline{3-3}
  Inv. Pen. 4 & & 143, 42 & 0.04 \\
  \hline
  Mountain Car 1 & \multirow{4}{*}{2, 50, 190, 3} & 406, 117 & 0.06 \\
  \cline{1-1}
  \cline{3-3}
  Mountain Car 2 & & 404, 115 & 0.04 \\
  \cline{1-1}
  \cline{3-3}
  Mountain Car 3 & & 404, 117 & 0.02 \\
  \cline{1-1}
  \cline{3-3}
  Mountain Car 4 & & 407, 118 & 0.04 \\
  \hline
  Pendulum 1 & \multirow{4}{*}{3, 32, 32, 32, 1} & 154, 41 & 0.05 \\
  \cline{1-1}
  \cline{3-3}
  Pendulum 2 & & 154, 41 & 0.02 \\
  \cline{1-1}
  \cline{3-3}
  Pendulum 3 & & 154, 41 & 0.02 \\
  \cline{1-1}
  \cline{3-3}
  Pendulum 4 & & 161, 48 & 0.65 \\
  \hline
  Acrobot 1 & \multirow{4}{*}{6, 64, 168, 3} & 579, 162 & 0.48 (3 problems) \\
  \cline{1-1}
  \cline{3-3}
  Acrobot 2 & & 569, 152 & 0.66 (3 problems) \\
  \cline{1-1}
  \cline{3-3}
  Acrobot 3 & & 590, 173 & 3.31 \\
  \cline{1-1}
  \cline{3-3}
  Acrobot 4 & & 609, 192 & 47.28 \\
  \hline
  Reuters & 1000, 512, 46 & 1546, 526 & 40.30 \\
  \hline
  MNIST & 4608, 128, 10 & 5002, 128 & 8.54 \\
  \hline
\end{tabular}
\label{table:bench}
\caption{Experimental results for the three neural networks
  described. Vars column refers to the number of variables in the LP:
  both continuous and binary.}
\end{table}

{\bf Inverted pendulum.} The neural network's description for this
scenario, including its size and architecture, is described in
Section~\ref{sec:experiments}. As reported in Table~\ref{table:bench},
the method here proposed solved each reachability query in less than 1
sec. We checked several other reachability specifications, here not
reported, to evaluate the performance degradation as a function of the
specification. We could not find specifications that could not be
solved in under 1 sec. We conclude that the methodology presented can
be used to evaluate any reachability problem for the IPCP controller
as here synthesised.

{\bf Acrobot, Pendulum and Mountain Car.} These are well-known
problems from classic control
theory~\cite{sutton+96a,furuta+91a,moore+90b}. As with the inverted
pendulum problem, we trained agent networks which solve these problems
and evaluated specifications using our toolkit. For the acrobot and
pendulum problems, the networks rely on non-linear trigonometric
functions; we generate a piece-wise linear approximation for these.

The specifications verified in the Pendulum and Mountain Car
benchmarks but with networks which are both larger and have
different layer sizes.

Similarly to the case above, the time taken to perform verification
was less than 1 sec. Differently from the pendulum case, the
specifications verified for the acrobot problem are much more complex
(mirroring the problem itself); consequently the specifications took
up to 45 sec to be verified. The third and fourth specifications for
acrobot are especially demanding as they require a full search of a
large part of a relatively high dimension state space to find a
solution. Taking this into account, we believe the performance to be
more than acceptable.

{\bf Reuters text classification.} This neural network is intended to
solve the problem of classifying articles from their
content~\cite{Reuters21578}. The network is composed of a binary input
layer followed by a hidden layer with a ReLU activation function
followed by the output layer. The structure of network is thus rather
shallow, but contains a large number of input and hidden neurons.

To evaluate the performance of the approach, we fixed all but 50 of
the inputs to known values and attempted to carry out reachability
analysis on the remaining inputs. We also neglected the use of the
softmax function to reduce the complexity resulting from its use.

We were able to replicate the existence of an input for a
(pre-softmax) output of the network for which we knew that a binary
input existed. We were able to solve the LP problem and find the
corresponding values in just under 45 secs. As above, considering the
size of the hidden layers and the number of binary variables present
in the corresponding problem, we find the performance to be attractive.

{\bf MNIST Image Recognition.} This neural network was put forward to
perform image recognition on the MNSIT handwritten numeric digits
dataset~\cite{lecun+10a}. The network is composed of a convolutional
part with several filters and a max-pooling layer followed by a hidden
feed-forward layer and an output layer.

We only considered the hidden and output layers of the network as
these are the feed-forward portion of the network. As in the previous
example the size of these layers is between $10^2$ and $10^3$
nodes. As in the previous experiment we attempted to find an input for
some output for which we know an input exists. As before, we did not
use softmax function when encoding the output.

The toolkit was able to find the input in just over 2 secs. Given the
size of the network and the number of variables in the corresponding
problem, we again evaluate the performance positively. Comparing the
last two scenarios, we conjecture that the large increase of binary
variables in the problem caused by the binary constraints on the input
creates the large performance gap between the Reuters dataset and
MNIST.

In summary the results suggest that the methodology developed, when
paired with the optimisations here studied, can solve the reachability
problem for several neural networks of interest. In particular we were
able to solve reachability analysis for deep nets of 3 layers of
significant size. The experiments demonstrate that performance depends
on a number of factors the most important being the size of the state
space searched, the number of variables (especially binary variables),
and the number of constraints.

\section{Conclusions}\label{sec:conclusions}
In this paper we have observed that while there is increasing
awareness that future society critical AI systems will need to be
verifiable, all of the major verification techniques fail to target
neural networks. Yet, neural networks are presently forecast to drive
most future AI systems.  We have attempted to begin to fill this gap
by providing a methodology for studying reachability analysis in feed
forward neural networks. Specifically, we have drawn a formal
correspondence between reachability in a neural network and an
associated linear programming problem. We have presented how to
circumvent problems caused by floating point arithmetic and optimise
the corresponding linear programming problem. The experimental results
shown demonstrate that the method can solve reachability for networks
of significant size.

Much work remains to do in this direction. We intend to study
recurrent networks and develop alternative techniques to solve the
reachability problem in recurrent networks. Also we intend to apply
the results of this work to synthesised controllers in engineering.

\small
\bibliography{vas,paper}

\end{document}